\documentclass{article}

\PassOptionsToPackage{}{natbib}

    \usepackage[preprint]{neurips_2023}

\usepackage[utf8]{inputenc} 
\usepackage[T1]{fontenc}    
\usepackage{hyperref}       
\usepackage{url}            
\usepackage{booktabs}       
\usepackage{amsfonts}       
\usepackage{nicefrac}       
\usepackage{microtype}      
\usepackage{xcolor}         

\usepackage{graphicx}
\graphicspath{ {./img/} }
\usepackage{subfigure}

\usepackage{amsmath}
\usepackage{bbm}

\usepackage {adjustbox}
\usepackage{multirow}

\newcommand\R{\mathbb{R}}

\newcommand\tmask{\texttt{[mask]}}
\newcommand\cls{\mathtt{[cls]}}
\newcommand\tcls{\texttt{[cls]}}

\title{Rethinking Data Augmentation for Tabular Data in Deep Learning}

\author{%
  Soma Onishi \\
  Meiji University\\
  \texttt{somaonishi4@gmail.com} \\
  \texttt{ce225005@meiji.ac.jp} \\
  \And
  Shoya Megro \\
  Meiji University\\
  \texttt{shoya.meguro28729@gmail.com} \\
  \texttt{ce235041@meiji.ac.jp} \\
}

\begin{document}
\maketitle
\begin{abstract}

Tabular data is the most widely used data format in machine learning (ML). While tree-based methods outperform DL-based methods in supervised learning, recent literature reports that self-supervised learning with Transformer-based models outperforms tree-based methods.
In the existing literature on self-supervised learning for tabular data, contrastive learning is the predominant method. In contrastive learning, data augmentation is important to generate different views.
However, data augmentation for tabular data has been difficult due to the unique structure and high complexity of tabular data.
In addition, three main components are proposed together in existing methods: model structure, self-supervised learning methods, and data augmentation. Therefore, previous works have compared the performance without comprehensively considering these components, and it is not clear how each component affects the actual performance.

In this study, we focus on data augmentation to address these issues.
We propose a novel data augmentation method, \textbf{M}ask \textbf{T}oken \textbf{R}eplacement (\texttt{MTR}), which replaces the mask token with a portion of each tokenized column; \texttt{MTR} takes advantage of the properties of Transformer, which is becoming the predominant DL-based architecture for tabular data, to perform data augmentation for each column embedding.
Through experiments with 13 diverse public datasets in both supervised and self-supervised learning scenarios, we show that \texttt{MTR} achieves competitive performance against existing data augmentation methods and improves model performance. In addition, we discuss specific scenarios in which \texttt{MTR} is most effective and identify the scope of its application. The code is available at \url{https://github.com/somaonishi/MTR/}.

\end{abstract}
\section{Introduction}
Deep learning is a powerful approach to solving a wide range of machine learning problems, such as image classification, natural language processing, and speech recognition. In recent years, deep learning has achieved impressive results in these areas. In many cases, it outperforms traditional machine learning approaches. For tabular data, however, tree-based approaches such as XGBoost~\citep{xgboost} and LigthGBM~\citep{lightgbm} dominate.

Although tree-based methods have achieved significant results in many applications, it has recently been reported that deep learning models, especially Transformer-based models, sometimes outperform tree-based methods~\citep{fttrans,somepalli2021saint,wang2022transtab,excelformer}.
However, \citet{whydotree} and \citet{shwartz2022tabular} argue that these reports are insufficiently validated and that tree-based methods are still dominant.
Since there are no established benchmarks for tabular data, such as ImageNet for computer vision or GLUE for NLP, this tree-based vs. deep learning trend is expected to continue for some time.

One of the main advantages of deep learning is the attractive ability to build multi-modal pipelines for problems where only part of the input is tabular and the other parts include image, text, audio, or other DL-friendly data. Such pipelines can then be trained end-to-end by gradient optimization for all modalities. This is especially useful for complex real-world problems where data comes from multiple sources and has different formats and structures. On the other hand, tree-based methods are limited to handling tabular data and cannot easily incorporate other types of data. In other words, improving the performance and robustness of deep learning on tabular data remains an important task.

Data augmentation is a powerful approach to further improve the performance and robustness of deep learning. Data augmentation has been widely used in image and language deep learning. However, data augmentation for tabular data is not well established. Due to the unique structure and high complexity of tabular data, it is difficult to apply existing data augmentation methods widely used for images and other types of data.

In recent years, self-supervised learning has also attracted the attention of researchers in the field of tabular data. In this field, contrastive learning such as SimCLR~\citep{chen2020simple} has become mainstream, and data augmentation is considered important to achieve good performance. Therefore, new data augmentation adapted to the unique structure of tabular data is still needed to improve the performance of contrastive learning on tabular data.

We summarize the contributions of our paper as follows:
\begin{itemize}
    \item We summarize data augmentations on existing tabular data and evaluate their performance on 13 diverse public datasets.
    \item We introduce \texttt{MTR}, a novel data augmentation that takes advantage of Transformer's properties to be highly competitive against existing data augmentations.
    \item We discuss scenarios in which \texttt{MTR} is more effective than other methods.
\end{itemize}
\section{Related work}
Data augmentation is a technique used to improve the generalization performance and robustness of machine learning models.
In the image domain, common methods include rotation, flipping, and cropping. More complex methods include Random Erasing~\citep{zhong2020random}, Cutmix~\citep{Yun_2019_ICCV}, Mixup~\citep{zhang2017mixup}, Manifold Mixup~\citep{ manifoldmixup}, and various other methods have been proposed~\citep{shorten2019survey}.
In addition, for the NLP task, approaches such as masking, reordering, and deleting words have been proposed ~\citep{https://doi.org/10.48550/arxiv.2105.03075}.

Recently, data augmentation using generative models such as GAN-based~\citep{NIPS2014_5ca3e9b1} and diffusion-based~\citep{song2021denoising} have been proposed in the image domain~\citep{antoniou2018data,mariani2018bagan,rashid2019skin,bird2022fruit,trabucco2023effective,azizi2023synthetic,akrout2023diffusion,xiao2023multimodal}. In the tabular data domain, SMOTE~\citep{chawla2002smote} and GAN-based~\citep{schultz4332129convgen,engelmann2021conditional} methods have been proposed. Theoretically, they can be extended to data augmentation. On the other hand, applying the generative model to data augmentation involves additional computational costs that cannot be ignored.

Data augmentation in images is mostly proposed in the context of supervised learning, but in the case of tabular data, it is mostly proposed as an adjunct to self-supervised learning methods.
Existing works on tabular data include autoencoder based~\citep{doi:10.1126/science.1127647,10.1145/1390156.1390294}, masked modeling based~\citep{vime,majmundar2022met,onishi2023tabret} and contrastive learning based~\citep{scarf,ucar2021subtab,wang2022transtab,contrastivemixup,somepalli2021saint,hajiramezanali2022stab} have been proposed, and the recent mainstream is contrastive learning based.
VIME~\citep{vime} and SCARF~\citep{scarf} proposed a data augmentation that randomly masks the input data and replaces the masked values by sampling from the same column. Using this data augmentation, VIME introduced the tasks of reconstructing the original features and estimating the mask vector from the masked features into pre-training; SCARF performed contrast learning using two views of the original data and the corrupted data.
SubTab~\citep{ucar2021subtab} and Transtab~\citep{wang2022transtab} introduced data augmentation to divide the input tabular data into multiple subsets. They performed contrastive learning using the features of the divided subsets.
Contrastive Mixup~\citep{contrastivemixup} uses manifold mixup to create different views for contrastive learning.
SAINT~\citep{somepalli2021saint} uses cutmix and manifold mixup to create different views for contrastive learning.
Stab~\citep{hajiramezanali2022stab} creates different views by applying a sparse matrix to the weights and performs SimSiam-like~\citep{9578004} contrastive learning.

\citet{excelformer} proposed a novel Transformer structure for tabular data, introducing two types of data augmentation: FeatMix in the input layer and HiddenMix in the hidden layer. In addition, Excelformer and FeatMix perform transformations that take into account the amount of mutual information.

Thus, while self-supervised learning has been the focus of much research on tabular data, there has been insufficient research on data augmentation techniques.
\section{Method}
\begin{figure}[htbp]
  \centering
  \includegraphics[width=1.0\textwidth]{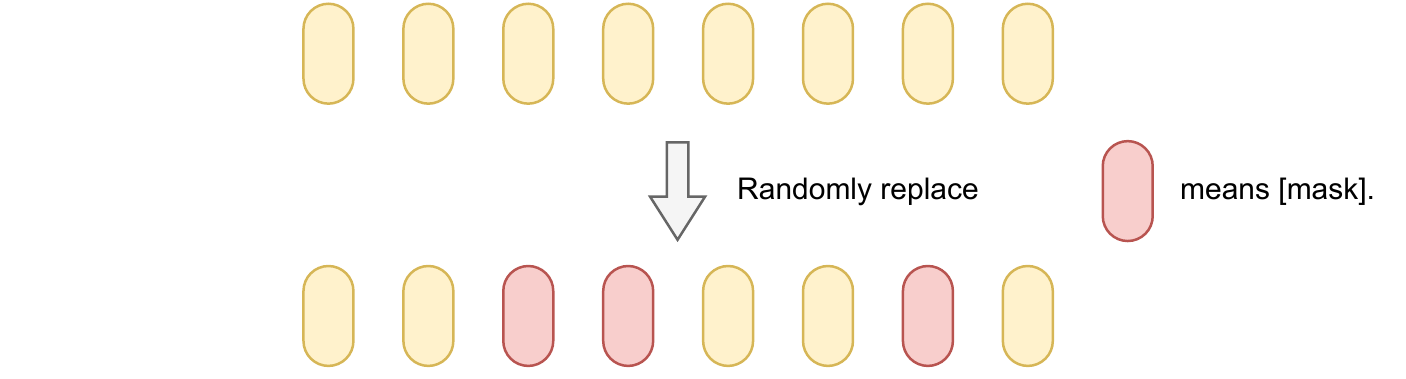}
  \caption{The demonstration of our method, \texttt{MTR}, a data augmentation method that randomly replaces part of the sequence of embedding tokenized by \texttt{Tokenizer} with \tmask (red token).}
  \label{fig:our_method}
\end{figure}

In this section, we introduce our proposed method. Our data augmentation methods can be easily implemented by extending the existing Transformer model.
First, given $k$ feature data $x \in \mathcal{X} \subseteq \R^k$, the Transformer model for tabular data transforms it into a sequence of embedding $T \in \R^{k \times d}$ using \texttt{Tokenizer}, where $d$ is the dimension of the embedding.
Then, using a \texttt{Transformer} such as vanilla-Transformer~\citep{vaswani2017attention}, \tcls token, and \texttt{Head} which predicts the target from encoded \tcls, the prediction is output from the sequence of embedding obtained as follows:

\begin{alignat*}{2}
    &T = \mathtt{Tokenizer}(x) && \in \R^{k \times d},\\
    &T' = \mathtt{Append}[T, \cls] && \in \R^{(k + 1) \times d},\\
    &T_{\mathrm{encode}} = \mathtt{Transformer}(T') && \in \R^{(k + 1) \times d},\\
    &\hat{y} = \mathtt{Head}(T_{\mathrm{encode}}^\cls).
\end{alignat*}

Our method focuses on a sequence of embedding $T$ generated by \texttt{Tokenizer}. Our method is simple: replace part of the sequence of embedding $T$ generated by \texttt{Tokenizer} with mask token (\tmask)~\citep{devlin-etal-2019-bert}. See Figure~\ref{fig:our_method}. We call our method \textbf{M}ask \textbf{T}oken \textbf{R}eplacement (\texttt{MTR}). Based on the Bernoulli distribution of the probability $p_m$, \texttt{MTR} decides whether to replace the column embedding with \tmask or not (1: replace, 0: keep). Each \tmask is a learnable parameter that is shared among the columns.

\section{Experiments}
In this section, we perform two validations to demonstrate the effectiveness of our method.

\begin{itemize}
    \item Experiment 1. Fully supervised with all \texttt{train}.
    \item Experiment 2. Self-supervised learning with 25\% of \texttt{train}.
\end{itemize}

\paragraph{Datasets.}
We use 13 diverse public datasets. For each dataset, we perform a split with $\mathtt{train} : \mathtt{val} : \mathtt{test} = 6 : 2 : 2$. The same split is performed for all methods, and a different split is performed for each seed. \texttt{val} is truncated to 20,000 samples for speed. 
The datasets include: Jasmine~\citep{Guyon2019} (JA), Spambase~\citep{Dua:2019} (SP), Philippine~\citep{Guyon2019} (PH), Wine Quality~\citep{CORTEZ2009547} (WQ), Gesture Phase Segmentation~\citep{Dua:2019} (GPS), Shrutime\footnote{\url{https://www.kaggle.com/datasets/shrutimechlearn/churn-modelling}} (SH), Online Shoppers\citep{online_shoppers} (OS), FIFA (FI), California Housing\footnote{\url{https://scikit-learn.org/stable/modules/generated/sklearn.datasets.fetch_california_housing.html}}~\citep{ca_housing} (CA), Bank Marketing~\citep{bank_marketing} (BM), Adult~\footnote{\url{https://www.kaggle.com/datasets/wenruliu/adult-income-dataset}}~\citep{kohavi-nbtree} (AD), Volkert~\citep{Guyon2019} (VO), Creditcard~\citep{creditcard} (CC). For the Wine Quality dataset, we follow the pointwise approach to learning-to-rank and treat ranking problems as regression problems. The dataset properties are summarized in Table~\ref{tab:dataset}.

\begin{table}[ht]
    \caption{Dataset properties. Notation: ``RMSE'' \textasciitilde\ root-mean-square error, ``ACC'' \textasciitilde\ accuracy, ``AUC'' \textasciitilde\ area under the ROC curve.}
    \label{tab:dataset}
    \centering
    \begin{tabular}{llrrll}
\toprule
Dataset & openml id & objects & features & metric & classes \\
\midrule
JA & 41143 & 2984 & 145 & AUC & 1 \\
SP & 44 & 4601 & 58 & AUC & 1 \\
PH & 41145 & 5832 & 309 & AUC & 1 \\
WQ & 287 & 6497 & 12 & RMSE & - \\
GPS & 4538 & 9873 & 33 & ACC & 5 \\
SH & 45062 & 10000 & 11 & AUC & 1 \\
OS & 45060 & 12330 & 18 & AUC & 1 \\
FI & 45012 & 19178 & 29 & RMSE & - \\
CA & - & 20640 & 9 & RMSE & - \\
BM & 1461 & 45211 & 17 & AUC & 1 \\
AD & - & 48842 & 15 & AUC & 1 \\
VO & 41166 & 58310 & 181 & ACC & 10 \\
CC & 1597 & 284807 & 30 & AUC & 1 \\
\bottomrule
\end{tabular}
\end{table}

\paragraph{Data preproccessing.}
The same preprocessing is applied to all datasets. We apply quantile transformation to numerical features and ordinal encoder from Scikit-learn library~\citep{pedregosa2011scikitlearn} to categorical features. We apply standardization to regression targets for all methods.

\paragraph{Model architecture.}
In this paper, we perform all experiments using FTTransformer~\citep{fttrans} as the base model for all methods. The hyperparameters of the model follow the default hyperparameters proposed by \citet{fttrans}. See our code for details.

\paragraph{Baseline.}
We include the following baselines for comparison:

\begin{itemize}
    \item \textbf{w/o data augmentation (w/o DA)}. Used to compare model performance without data augmentation.
    \item \textbf{Manifold Mixup}~\citep{manifoldmixup}.
    Apply in the latent space. The manifold mixup is applied to two $T_{\mathrm{encode}, i}^\cls, T_{\mathrm{encode}, j}^\cls$ in the same batch encoded by \texttt{Transformer} with $\lambda = \mathrm{Beta}(\alpha, \alpha)$. Label-mixing is performed as $y_i = (1-\lambda) y_i + \lambda y_j$. The hyperparameter is $\alpha$. See our code for more details.
    \item \textbf{Cutmix}.
    Applied in the input space. It is applied to two inputs $x_i, x_j \in \R^k$ in the same batch. Randomly select $\lfloor \lambda \times k \rfloor$ indices and swap the values of $x_i, x_j$ corresponding to the selected indices. Note that $\lambda \sim \mathrm{Beta}(\alpha, \alpha)$. Also, according to the actual swapped rate $\lambda'$, label-mixing is performed as $y_i = (1-\lambda') y_i + \lambda' y_j$. The hyperparameter is $\alpha$. See our code for more details.
    \item \textbf{SCARF}~\citep{scarf}.
    Applied in the input space. The mask vector $m$ generated from the Bernoulli distribution based on the mask fraction $p_m$ is used to determine the corruption points. When $m_j$ is 1, $x_j$ is replaced by $\hat{x}_j \sim \widehat{\mathcal{X}_j}$. Here $\widehat{\mathcal{X}_j}$ is assumed to be uniformly distributed over $\mathcal{X}_j = \{x_j : x \in \mathcal{X}\}$, where $\mathcal{X}$ is the training data. The hyperparameter is $p$. See our code for more details.
    \item \textbf{HiddenMix}~\citep{excelformer}.
    Applied in the latent space. For two $T_i, T_j \in \R^{k\times d}$ in the same batch generated by \texttt{Tokenizer}, $T_i = S \odot T_i + (\mathbbm{1} - S)\odot T_j $.
    The coefficient matrix $S$ and the all-one matrix $\mathbbm{1}$ are of size $k \times d$. $S = [s_1,s_2,\ldots, s_k]^\top$, all whose vector $s_h \in \R^d$ ($h=1,2,\ldots,k$) are identical and have $\lfloor \lambda \times d \rfloor$ randomly selected elements of 1's and the rest elements are 0's. Note that $\lambda \sim \mathrm{Beta}(\alpha, \alpha)$. label-mixing is performed as $y_i = \lambda y_i + (1 - \lambda) y_j$. The hyperparameter is $\alpha$. See our code for more details.
\end{itemize}

\paragraph{Supervised and self-supervised learning.}
In supervised learning, we applied data augmentation with a probability of 50\% for all methods and no data augmentation at test time.
In self-supervised learning, we perform pre-training with contrastive learning.
For the output $T_\mathrm{encode}^\cls$ of \texttt{Transformer}, we perform the transformation $z = \mathtt{PredictionHead}(T_\mathrm{encode}^\cls) \in \R^h$, where $h$ is the dimension of the latent space.
We applied this transformation to $\hat{T}_\mathrm{encode}^\cls$ with data augmentation and $T_\mathrm{encode}^\cls$ without data augmentation to obtain the vectors $\hat{z}, z$.
Then, we applied NT-Xent loss~\citep{chen2020simple} to these two different views $\hat{z}, z$ for contrast learning.
Note that Cutmix was not included in the baseline in the self-supervised learning because it does not use the label-mixing, and therefore behaves almost the same as SCARF.

\paragraph{Training.}
\begin{table}[t]
    \caption{Correspondence table between\texttt{train} size and batch size.}
    \label{tab:batch_size}
    \centering
    \begin{tabular}{lr}
\toprule
 train size &  batch size\\
 \midrule
 more than 50,000 & 1024\\
 between 10,000 and 50,000 & 512\\
 between 5,000 and 10,000 & 256\\
 between 1,000 and 5,000 & 128\\
 less than 1,000 & 64\\
 \bottomrule
\end{tabular}
\end{table}

We changed the batch size for all methods according to the size of the \texttt{train} (see Table~\ref{tab:batch_size}).
For supervised learning, we set the maximum training epochs to 500 for all methods. We also applied early stopping with patience of 15 using validation loss. For self-supervised learning, the early stopping patience was set to 10 and the maximum number of epochs was set to 200.
The hyperparameters $\alpha$ of Manifold Mixup, Cutmix, and HiddenMix were set to $\alpha = \{0.1, 0.2, 0.5, 0.75, 1.0, 1.5, 2.0\}$, and the hyperparameters $p_m$ of SCARF and \texttt{MTR} was set to $p_m = \{0.1, 0.2, 0.5, 2.0 0.1, 0.2, 0.3, 0.4, 0.5, 0.6, 0.7\}$.

In all of our experiments, we used the following two pieces of hardware; hardware 1 has an Intel(R) Xeon(R) Gold 6244 CPU with NVIDIA Quadro RTX 5000 GPU, and hardware 2 has an Intel(R) Xeon(R) Gold 6250 CPU with NVIDIA RTX A2000 12GB GPU. 

\subsection{Experiment 1. Supervised learning}
\label{sec:exp_1}
\begin{table}[t]
    \caption{Results for the supervised learning. The metrics are averaged over 10 random seeds. For each dataset, the top result is in \textbf{bold} and the second result is in \underline{underline}. For each dataset, ranks are computed by sorting the reported scores; the ``Rank'' column reports the average rank across all datasets.}
    \label{tab:experiment_1}
    \centering
    \begin{adjustbox}{max width=\textwidth}
\begin{tabular}{llllll}
\toprule
Dataset & JA \textuparrow & SP \textuparrow & PH \textuparrow & WQ \textdownarrow & GPS \textuparrow \\
\midrule
w/o DA  & $85.06$  & $98.53$  & $80.78$  & $0.6907$  & $56.45$  \\
Manifold Mixup ($\alpha$)  & $85.29$ (2.0)  & $98.48$ (0.1)  & $81.36$ (0.2)  & $\underline{0.6805}$ (0.75)  & $56.62$ (1.0)  \\
Cutmix ($\alpha$)  & $\underline{85.62}$ (1.5)  & $98.37$ (0.2)  & $81.53$ (1.0)  & $0.6814$ (0.2)  & $\underline{59.80}$ (1.0)  \\
SCARF ($p_m$)  & $85.30$ (0.4)  & $98.62$ (0.2)  & $81.41$ (0.7)  & $0.6819$ (0.2)  & $58.34$ (0.1)  \\
HiddenMix ($\alpha$)  & $85.38$ (2.0)  & $\mathbf{98.65}$ (0.5)  & $\underline{81.54}$ (1.0)  & $0.6809$ (1.5)  & $\mathbf{59.99}$ (0.75)  \\
\texttt{MTR} ($p_m$) & $\mathbf{85.63}$ (0.6) & $\underline{98.63}$ (0.7) & $\mathbf{82.91}$ (0.7) & $\mathbf{0.6804}$ (0.3) & $57.55$ (0.2) \\
\bottomrule
\end{tabular}
\end{adjustbox}

\vspace{0.4cm}

\begin{adjustbox}{max width=\textwidth}
\begin{tabular}{llllll}
\toprule
Dataset & SH \textuparrow & OS \textuparrow & FI \textdownarrow & CA \textdownarrow & BM \textuparrow \\
\midrule
w/o DA  & $86.18$  & $92.44$  & $10100.0$  & $0.4781$  & $93.63$  \\
Manifold Mixup ($\alpha$)  & $86.36$ (1.0)  & $\underline{92.55}$ (0.75)  & $9944.3$ (0.1)  & $0.4773$ (1.5)  & $93.64$ (2.0)  \\
Cutmix ($\alpha$)  & $\underline{86.46}$ (1.0)  & $92.37$ (0.2)  & $9814.5$ (0.5)  & $\underline{0.4691}$ (0.5)  & $93.64$ (0.1)  \\
SCARF ($p_m$)  & $\mathbf{86.46}$ (0.2)  & $92.43$ (0.1)  & $\mathbf{9690.1}$ (0.4)  & $0.4710$ (0.1)  & $93.67$ (0.1)  \\
HiddenMix ($\alpha$)  & $86.25$ (0.75)  & $\mathbf{92.62}$ (0.75)  & $9937.3$ (1.0)  & $0.4802$ (0.1)  & $\underline{93.72}$ (2.0)  \\
\texttt{MTR} ($p_m$) & $86.33$ (0.5) & $92.37$ (0.1) & $\underline{9779.8}$ (0.6) & $\mathbf{0.4675}$ (0.4) & $\mathbf{93.78}$ (0.6) \\
\bottomrule
\end{tabular}
\end{adjustbox}

\vspace{0.4cm}

\begin{adjustbox}{max width=\textwidth}
\begin{tabular}{llll|l}
\toprule
Dataset & AD \textuparrow & VO \textuparrow & CC \textuparrow & Rank \\
\midrule
w/o DA  & $91.73$  & $71.82$  & $97.83$  & $5.54$  \\
Manifold Mixup ($\alpha$)  & $91.75$ (0.2)  & $71.92$ (0.5)  & $98.26$ (1.0)  & $4.08$  \\
Cutmix ($\alpha$)  & $\underline{91.80}$ (2.0)  & $\mathbf{72.79}$ (0.5)  & $98.21$ (1.0)  & $3.08$  \\
SCARF ($p_m$)  & $91.78$ (0.3)  & $\underline{72.43}$ (0.3)  & $\mathbf{98.38}$ (0.3)  & $\underline{2.85}$  \\
HiddenMix ($\alpha$)  & $91.75$ (0.1)  & $72.27$ (0.75)  & $\underline{98.27}$ (2.0)  & $2.92$  \\
\texttt{MTR} ($p_m$) & $\mathbf{91.81}$ (0.5) & $72.06$ (0.1) & $98.20$ (0.4) & $\mathbf{2.54}$ \\
\bottomrule
\end{tabular}
\end{adjustbox}
\end{table}

\begin{figure}[t]
    \centering
    \includegraphics[width=\textwidth]{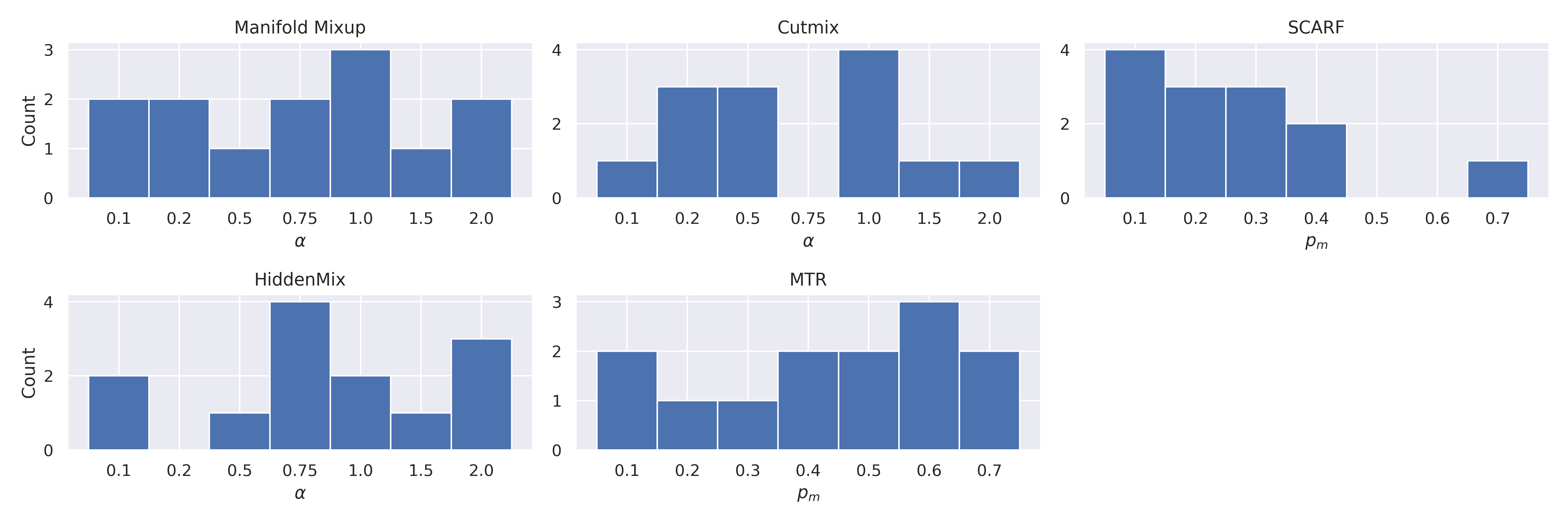}
    \caption{Comparison of hyperparameter distributions for different data augmentation methods in supervised learning. The histograms show the aggregated values of the hyperparameters for each of the data augmentation methods for all datasets.}
    \label{fig:hypara_hist}
\end{figure}

Table~\ref{tab:experiment_1} shows the results of supervised learning and Figure~\ref{fig:hypara_hist} shows the distribution of the selected best hyperparameters.
Overall, the data augmentation methods improve the performance compared to the case without data augmentation (w/o DA).
The proposed method, \texttt{MTR}, achieved competitive performance compared to other data augmentation methods. It achieved the highest average rank of 2.54, indicating that it is effective in improving model performance. In particular, it outperformed the other methods on PH dataset which has a large number of features.

SCARF, HiddenMix, and Cutmix were highly competitive with our methods. 
For SCARF, \citet{scarf} claimed that a default $p_m$ between 0.5 and 0.7 is better, but we found that a lower $p_m$ is better in many cases in this experiment.
HiddenMix and Cutmix significantly outperformed the proposed method on GPS and VO datasets for multi-class classification. This is because the label-mixing increases the diversity of the data. In fact, we confirmed that the accuracy of HiddenMix without label-mixing on GPS dataset was significantly lower -- 57.99\% (w/o label-mixing).

For Manifold Mixup, the overall performance was similar to that of w/o DA. The average rank was also lower than for other data augmentation methods.

\subsection{Experiment 2. Self-supervised learning}
\begin{table}[t]
    \caption{Results for the self-supervised learning. The metrics are averaged over 10 random seeds. For each dataset, the top result is in \textbf{bold} and the second result is in \underline{underline}. For each dataset, ranks are computed by sorting the reported scores; the ``Rank'' column reports the average rank across all datasets.}
    \label{tab:experiment_2}
    \centering
    \begin{adjustbox}{max width=\textwidth}
\begin{tabular}{llllll}
\toprule
Dataset & JA \textuparrow & SP \textuparrow & PH \textuparrow & WQ \textdownarrow & GPS \textuparrow \\
\midrule
Supervised  & $83.64$  & $\mathbf{97.81}$  & $76.70$  & $0.7178$  & $49.31$  \\
Manifold Mixup ($\alpha$)  & $83.72$ (2.0)  & $97.71$ (2.0)  & $76.53$ (0.75)  & $0.7174$ (1.5)  & $49.41$ (2.0)  \\
SCARF ($p_m$)  & $\mathbf{84.20}$ (0.6)  & $97.61$ (0.6)  & $\underline{79.29}$ (0.7)  & $0.7153$ (0.1)  & $49.37$ (0.3)  \\
HiddenMix ($\alpha$)  & $83.78$ (1.0)  & $\underline{97.74}$ (1.5)  & $76.85$ (2.0)  & $\mathbf{0.7138}$ (0.75)  & $\mathbf{49.81}$ (0.5)  \\
\texttt{MTR} ($p_m$) & $\underline{83.83}$ (0.5) & $97.45$ (0.5) & $\mathbf{79.70}$ (0.6) & $\underline{0.7152}$ (0.1) & $\underline{49.46}$ (0.2) \\
\bottomrule
\end{tabular}
\end{adjustbox}

\vspace{0.4cm}

\begin{adjustbox}{max width=\textwidth}
\begin{tabular}{llllll}
\toprule
Dataset & SH \textuparrow & OS \textuparrow & FI \textdownarrow & CA \textdownarrow & BM \textuparrow \\
\midrule
Supervised  & $\underline{84.33}$  & $91.60$  & $\mathbf{11094.9}$  & $0.5378$  & $92.16$  \\
Manifold Mixup ($\alpha$)  & $84.07$ (0.75)  & $91.35$ (1.5)  & $11328.8$ (0.75)  & $0.5437$ (0.75)  & $92.06$ (0.1)  \\
SCARF ($p_m$)  & $84.09$ (0.7)  & $\mathbf{91.88}$ (0.5)  & $11410.9$ (0.2)  & $\underline{0.5266}$ (0.3)  & $\underline{92.21}$ (0.2)  \\
HiddenMix ($\alpha$)  & $\mathbf{84.35}$ (0.5)  & $91.58$ (0.1)  & $\underline{11269.8}$ (1.5)  & $0.5408$ (1.5)  & $92.05$ (0.2)  \\
\texttt{MTR} ($p_m$) & $84.24$ (0.7) & $\underline{91.74}$ (0.3) & $11345.4$ (0.4) & $\mathbf{0.5237}$ (0.5) & $\mathbf{92.31}$ (0.3) \\
\bottomrule
\end{tabular}
\end{adjustbox}

\vspace{0.4cm}

\begin{adjustbox}{max width=\textwidth}
\begin{tabular}{llll|l}
\toprule
Dataset & AD \textuparrow & VO \textuparrow & CC \textuparrow & Rank \\
\midrule
Supervised  & $91.32$  & $\mathbf{63.44}$  & $96.10$  & 3.23  \\
Manifold Mixup ($\alpha$)  & $91.30$ (0.5)  & $63.04$ (0.5)  & $\underline{97.14}$ (2.0)  & $3.92$  \\
SCARF ($p_m$)  & $91.32$ (0.6)  & $\underline{63.41}$ (0.7)  & $\mathbf{97.37}$ (0.1)  & $\underline{2.62}$  \\
HiddenMix ($\alpha$)  & $\underline{91.35}$ (0.5)  & $62.82$ (1.5)  & $97.11$ (0.1)  & $2.69$  \\
\texttt{MTR} ($p_m$) & $\mathbf{91.38}$ (0.6) & $62.15$ (0.4) & $97.11$ (0.4) & $\mathbf{2.54}$ \\
\bottomrule
\end{tabular}
\end{adjustbox}
\end{table}

\begin{figure}[t]
    \centering
    \includegraphics[width=\textwidth]{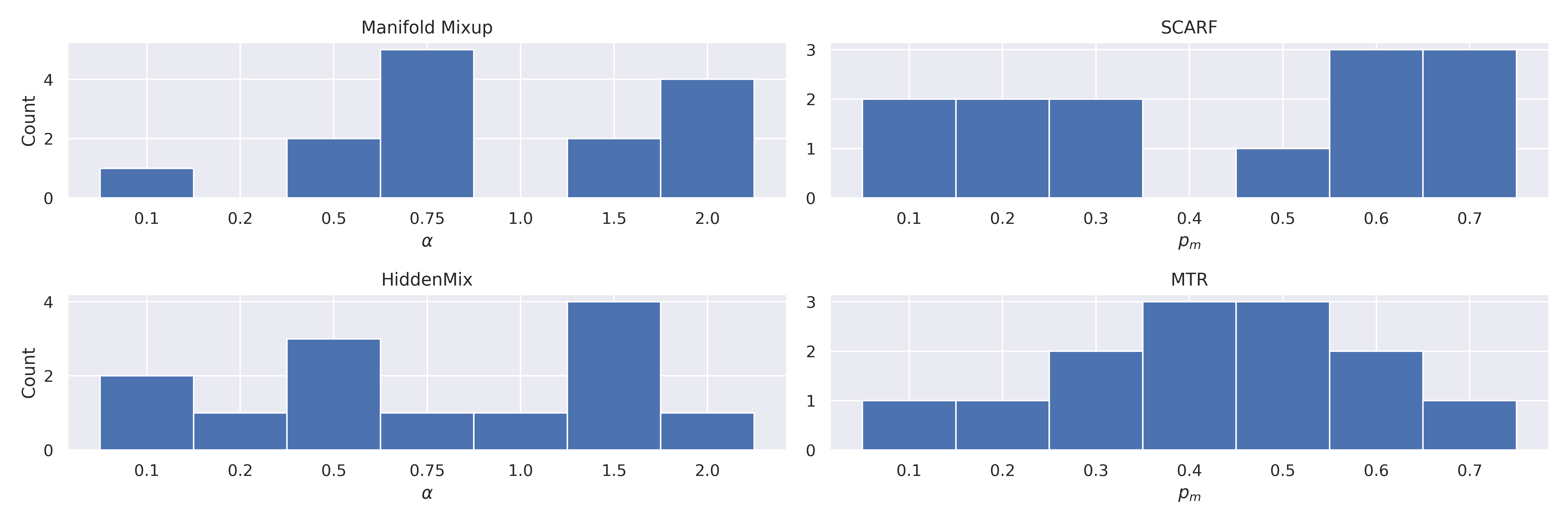}
    \caption{Comparison of hyperparameter distributions for different data augmentation methods in self-supervised learning. The histograms show the aggregated values of the hyperparameters for each of the data augmentation methods for all datasets.}
    \label{fig:hypara_hist_self}
\end{figure}

Table~\ref{tab:experiment_2} shows the results of the self-supervised learning, and Figure~\ref{fig:hypara_hist} shows the distribution of the selected best hyperparameters.
Overall, \texttt{MTR} performed the best based on the mean ranks. In particular, it outperformed the other methods on PH dataset.
SCARF and HiddenMix also performed better than supervised methods overall, confirming their competitiveness against \texttt{MTR}.
Self-supervised learning with Manifold Mixup had the only negative effect.
For most of the datasets, the pre-training benefit from self-supervised learning was marginal.

\subsection{Ablation study}

\begin{table}[t]
    \caption{Results for the ablation study in supervised learning. The metrics are averaged over 10 random seeds. For each dataset, the better score is in \textbf{bold}. For each dataset, ranks are computed by sorting the reported scores; the ``Rank'' column reports the average rank across all datasets.}
    \label{tab:ablation_study}
    \centering
    \begin{adjustbox}{max width=\textwidth}
\begin{tabular}{llllll}
\toprule
Dataset & JA \textuparrow & SP \textuparrow & PH \textuparrow & WQ \textdownarrow & GPS \textuparrow \\
\midrule
After bias ($p_m$)  & $\mathbf{85.63}$ (0.6)  & $\mathbf{98.63}$ (0.7)  & $\mathbf{82.91}$ (0.7)  & $\mathbf{0.6804}$ (0.3)  & $57.55$ (0.2)  \\
Before bias ($p_m$) & $85.50$ (0.7) & $98.61$ (0.2) & $82.72$ (0.7) & $0.6833$ (0.3) & $\mathbf{58.34}$ (0.1) \\
\bottomrule
\end{tabular}
\end{adjustbox}

\vspace{0.4cm}

\begin{adjustbox}{max width=\textwidth}
\begin{tabular}{llllll}
\toprule
Dataset & SH \textuparrow & OS \textuparrow & FI \textdownarrow & CA \textdownarrow & BM \textuparrow \\
\midrule
After bias ($p_m$)  & $\mathbf{86.33}$ (0.5)  & $92.37$ (0.1)  & $9779.8$ (0.6)  & $\mathbf{0.4675}$ (0.4)  & $\mathbf{93.78}$ (0.6)  \\
Before bias ($p_m$) & $86.31$ (0.4) & $\mathbf{92.39}$ (0.1) & $\mathbf{9746.2}$ (0.7) & $0.4715$ (0.1) & $93.75$ (0.3) \\
\bottomrule
\end{tabular}
\end{adjustbox}

\vspace{0.4cm}

\begin{adjustbox}{max width=\textwidth}
\begin{tabular}{lllll}
\toprule
Dataset & AD \textuparrow & VO \textuparrow & CC \textuparrow & Rank \\
\midrule
After bias ($p_m$)  & $91.81$ (0.5)  & $72.06$ (0.1)  & $98.20$ (0.4)  & 1.46  \\
Before bias ($p_m$) & $\mathbf{91.83}$ (0.4) & $\mathbf{72.20}$ (0.5) & $\mathbf{98.41}$ (0.5) & 1.54 \\
\bottomrule
\end{tabular}
\end{adjustbox}

\end{table}

In this section, we test design choices of \texttt{MTR}.

\texttt{Tokenizer} in tabular data, including the FT-Transformer's \texttt{Tokenizer} (\citet{fttrans} call FeatureTokenizer) used in this experiment, generally consists of weights $W \in \R^d$ and biases $b \in \R^d$.
For example, FeatureTokenizer is defined for the j-th column using \mbox{$W^{(num)}_j \in \R^d$} for the numeric column and lookup table \mbox{$W^{(cat)}_j \in \R^{S_{j} \times d}$} for the category column as follows:

\begin{alignat*}{2}
    &T^{(num)}_j  = b^{(num)}_j + x^{(num)}_j \cdot W^{(num)}_j && \in \R^d, \\
    &T^{(cat)}_j  = b^{(cat)}_j  + e_j^T W^{(cat)}_j             && \in \R^d, \\
    &T            = \mathtt{stack} \left[ T^{(num)}_1,\ \ldots,\ T^{(num)}_{k^{(num)}},\ T^{(cat)}_1,\ \ldots,\ T^{(cat)}_{k^{(cat)}} \right]              && \in \R^{k \times d}.
\end{alignat*}
where $e_j^T$ is a one-hot vector for the corresponding categorical feature.

We can consider the weights $W$ as directly representing the column value, and the bias $b$ as representing the column itself.
Here, our question is whether it is better to apply \texttt{MTR} after bias or before bias. After bias (normal \texttt{MTR}), the subsequent network is not able to recognize the column itself. On the other hand, if \texttt{MTR} is applied before bias, only the column value is masked, and the subsequent network is able to recognize the masked column itself.

Table~\ref{tab:ablation_study} shows the results of the ablation study in supervised learning. The results are almost the same whether \texttt{MTR} is applied after bias or before bias.
Since applying \texttt{MTR} before bias requires modification of the existing \texttt{Tokenizer}, we conclude that applying \texttt{MTR} after bias is superior in terms of implementation cost.

\section{Discussion}
\subsection{When is \texttt{MTR} better than the other data augmentations?}
\label{sec:when_mtr_is_better}
\begin{figure}[t]
    \centering
    \includegraphics[width=\textwidth]{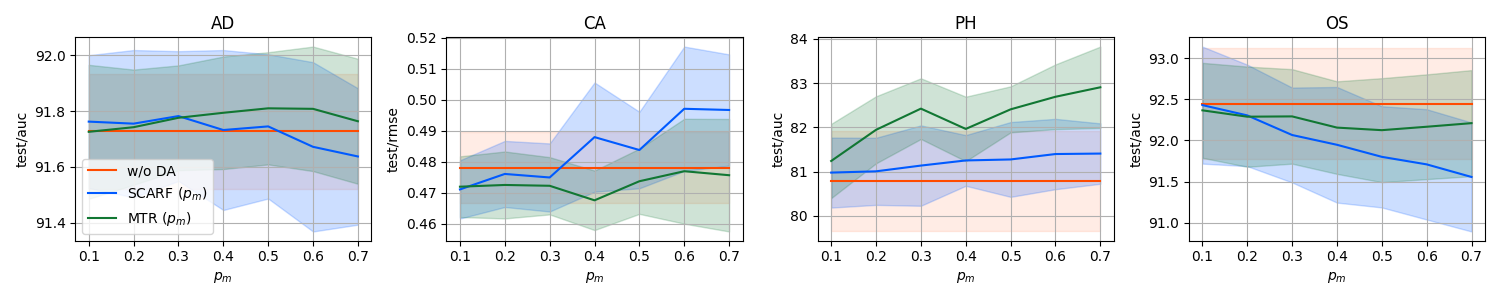}
    \caption{Comparison of SCARF and \texttt{MTR} accuracy variation with mask ratio: SCARF shows significant degradation when a high mask ratio is not suitable, while \texttt{MTR} shows relatively low degradation.}
    \label{fig:scarf_vs_mtr}
\end{figure}
In this section, we discuss scenarios in which \texttt{MTR} outperforms other data augmentation methods.

From Experiment 1, we conclude that \texttt{MTR} is effective for datasets with a large number of features, such as the PH dataset. Such datasets are likely to have high redundancy between columns; \texttt{MTR} hides the information and thus performs well for highly redundant datasets. However, for data with high dependency between columns, it may have a negative impact because necessary information may be lost. For example, in Table~\ref{tab:experiment_1}, we can infer that the value of $p_m$ is not suitable for OS and VO datasets with a minimum value of 0.1.

SCARF is sensitive to hyperparameters. Figure~\ref{fig:scarf_vs_mtr} compares \texttt{MTR} and SCARF on several datasets; SCARF can be negatively affected by unsuitable hyperparameters, while \texttt{MTR} shows little degradation even at high values of $p_m$.

Mixup-family data augmentation requires label-mixing. Multi-modal learning, especially when there are no labels, such as generative models, makes it impossible to use these data augmentations because label-mixing is not applicable.

Based on the above, we conclude that \texttt{MTR} is more suitable than other methods for the following scenarios:
\begin{itemize}
    \item Training on datasets with redundant columns.
    \item Training on multi-modal pipelines.
\end{itemize}

\subsection{\texttt{MTR} is like Random Erasing.}
Random Erasing~\citep{zhong2020random}, data augmentation in the image domain, replaces the pixel value of any patch in the image with a random value from [0, 255]; in the case of ViT~\citep{dosovitskiy2021an}, the embedding of a patch replaced by a random value is an irrelevant embedding of the other patches.
In the case of \texttt{MTR}, any sequence of embedding is replaced by \tmask, which is a learnable parameter shared by all columns and thus becomes a vector unrelated to the unmasked columns.
In other words, while Random Erasing in ViT generates irrelevant embedding from meaningless patch, \texttt{MTR} directly replaces them with irrelevant vector \tmask.
Thus, \texttt{MTR} can be considered as Random Erasing for tabular data.

\subsection{Limitations}
Due to the property of \texttt{MTR} to mask information between columns, it may have a negative impact on data with high dependency between columns, since the necessary information may be lost. For example, in Table~\ref{tab:experiment_1}, we can infer that the value of $p_m$ is not suitable for the OS and VO datasets with a minimum value of 0.1.
In other words, it is likely to not work well for data with a high degree of independence among columns and label values that depend on information in almost all columns.
However, using an indicator of the relationship between columns, such as the mutual information content used by \citet{excelformer}, and setting a sort of priority level for the columns to be masked, this negative effect is expected to be reduced.

As we concluded in Section~\ref{sec:when_mtr_is_better}, \texttt{MTR} is likely to perform relatively well on highly redundant datasets. Highly redundant data is likely to have a large number of columns, but models using the Transformer require more resources for training when the number of columns is large. The main cause of this issue is that the complexity of the vanilla-MHSA (Multi-Head Self-Attention) is quadratic with respect to the number of columns. However, this issue can be alleviated by using an efficient approximation of the MHSA~\citep{tay2022efficient}.

\section{Conclusion}
This study investigates the status quo in the field of data augmentation for tabular data, which has been important but has not received much attention until now. We performed a fair comparison of each data augmentation method using the same network.
We also proposed a novel data augmentation method, \texttt{MTR}, which focuses on the sequence of embedding in Transformer, and showed that it is highly competitive with other data augmentation methods.
Furthermore, based on experimental results and the properties of \texttt{MTR}, we provided a scenario in which \texttt{MTR} outperforms other methods.
We hope that our work will provide further developments in deep learning for tabular data.

\bibliography{ref}
\bibliographystyle{plainnat}

\newpage
\appendix
\renewcommand\thesubsection{\Alph{subsection}}

\section*{Supplementary material}

\subsection{Hyperparameters of FTTransformer}
We used the default hyperparameters of FTTransformer proposed by \citet{fttrans} in all experiments. The details of the hyperparameters are shown in Table~\ref{tab:fttrans-config}.

\begin{table}[h]
    \caption{Default FTTransformer used in the main text. ``FFN size factor'' is a ratio of the \texttt{FFN}'s hidden size to the feature embedding size. Note that the weight decay is 0.0 for \texttt{FeatureTokenizer}, \texttt{LayerNorm}, and \texttt{bias}es.}
    \label{tab:fttrans-config}
    \centering
    \vspace{1em}
    \begin{tabular}{lc}
        \toprule
        Layer count &  3\\
        Feature embedding size & 192\\
        Head count & 8 \\
        Activation \& FFN size factor & (\texttt{ReGLU}, $\nicefrac{4}{3}$) \\
        Attention dropout & 0.2 \\
        FFN dropout & 0.1 \\
        Residual dropout & 0.0 \\
        Initialization & \texttt{Kaiming}\citep{he2015delving} \\
        \midrule
        Optimizer & \texttt{AdamW} \\
        Learning rate & 1e-4 \\
        Weight decay & 1e-5 \\
        \bottomrule
    \end{tabular}
\end{table}

\subsection{Results for all hyperparameters of the experiment}
\begin{figure}
    \centering
    \includegraphics[width=\textwidth]{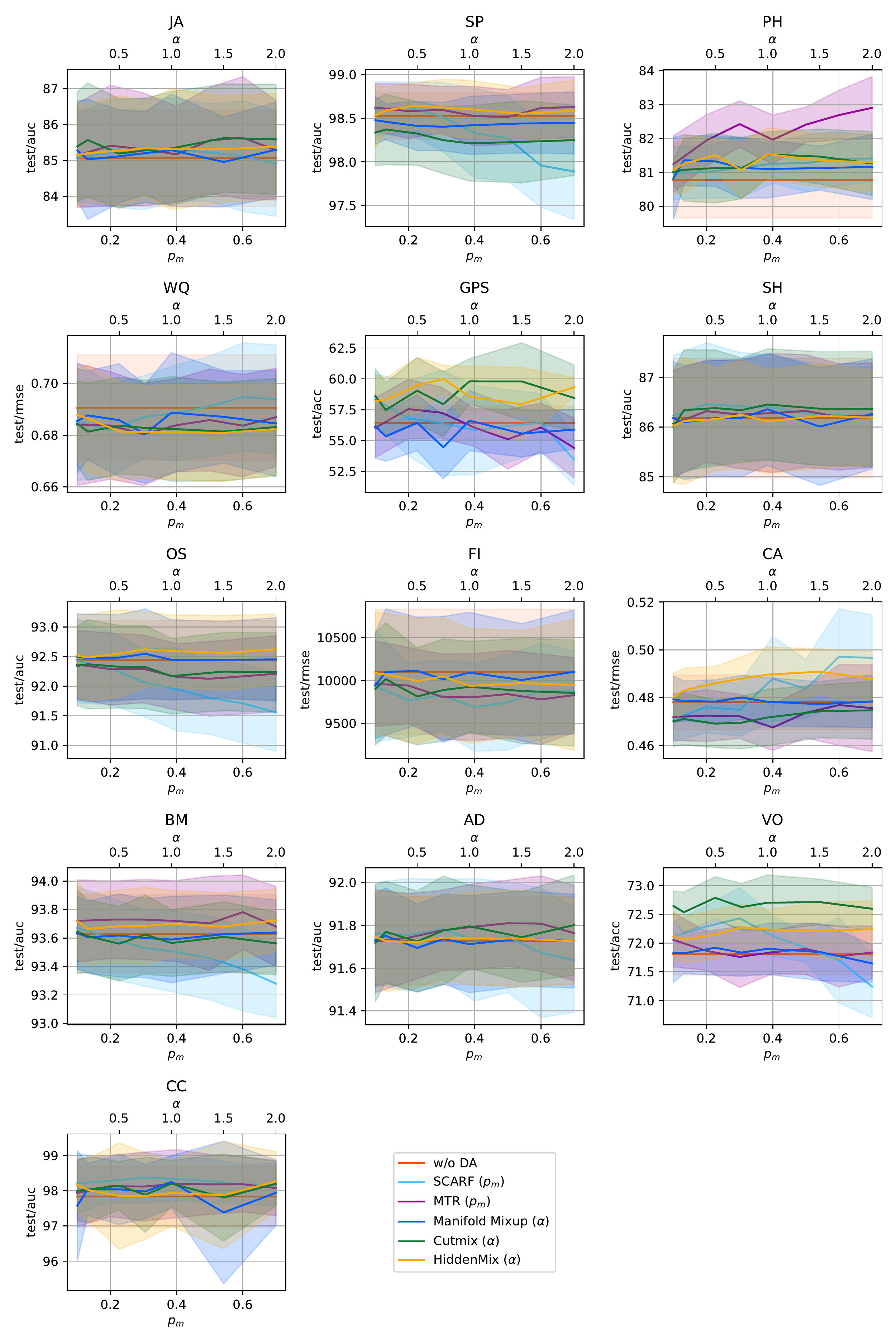}
    \caption{Results for all hyperparameters in Experiment 1. The solid line represents the average score of each method, while the translucent interval indicates the range of standard deviation $\pm\sigma$.}
    \label{fig:all_results_exp_1}
\end{figure}
\begin{figure}
    \centering
    \includegraphics[width=\textwidth]{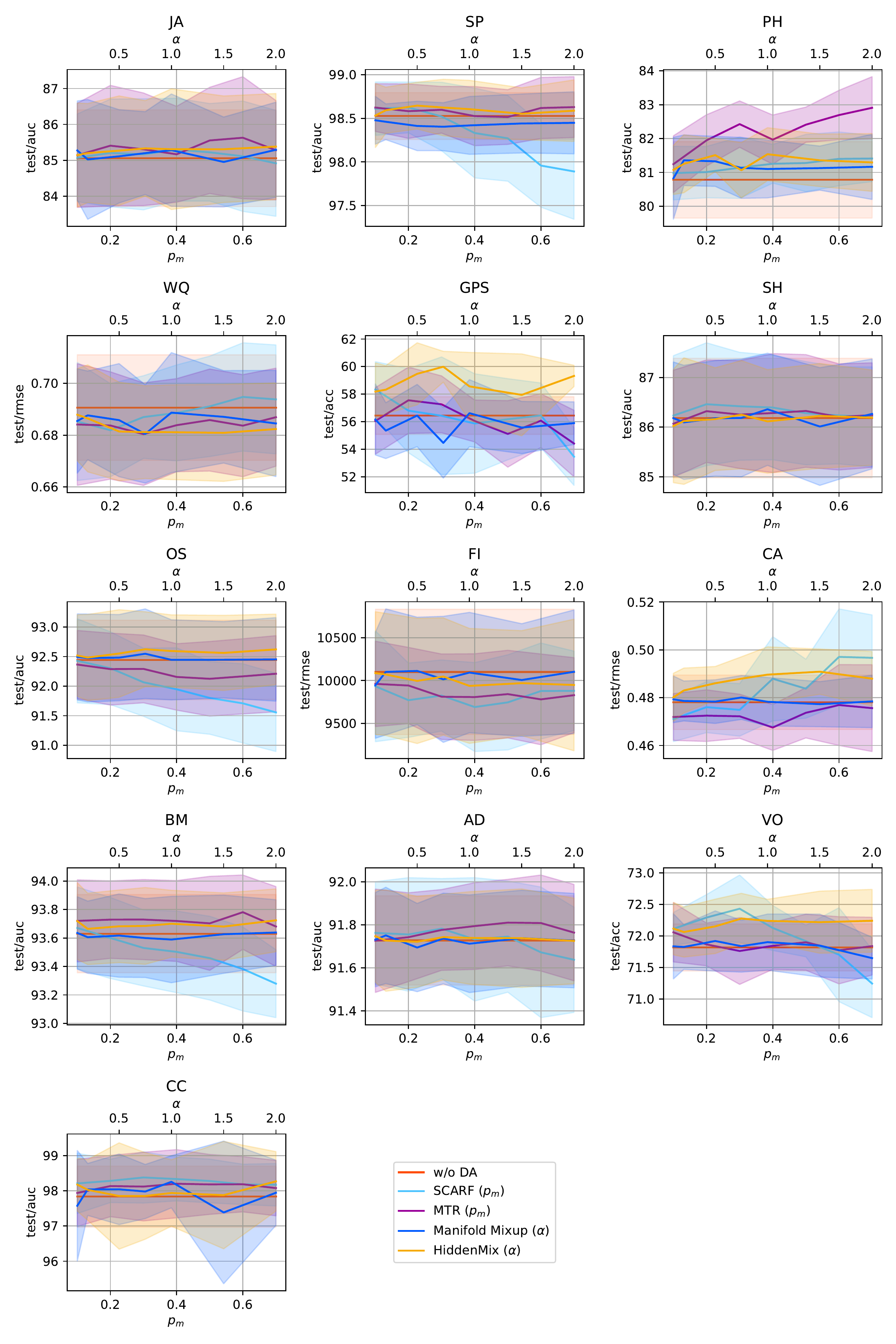}
    \caption{Results for all hyperparameters in Experiment 2. The solid line represents the average score of each method, while the translucent interval indicates the range of standard deviation $\pm\sigma$.}
    \label{fig:all_results_exp_2}
\end{figure}

Although we reported only the best hyperparameters in the main text, we provide all the results in this section. The results for all hyperparameters in Experiment 1 are shown in Figure~\ref{fig:all_results_exp_1}, and the results for all hyperparameters in Experiment 2 are shown in Figure~\ref{fig:all_results_exp_2}.

\end{document}